\documentclass[11pt]{article}

\usepackage[preprint]{acl}

\usepackage{times}
\usepackage{latexsym}

\usepackage[T1]{fontenc}

\usepackage[utf8]{inputenc}

\usepackage{microtype}

\usepackage{inconsolata}

\usepackage{graphicx}

\title{
Enhancing Generative Information Extraction with Two-step Validation:\\A Product Attribute Use Case
}

\author{Yi-Sheng Hsu \qquad Nermeen Abou Baker \qquad Uwe Handmann \\
Computer Science Institute, Ruhr West University of Applied Sciences \\
Bottrop, Germany \\
\texttt{\{firstname.lastname\}@hs-ruhrwest.de}
}

\usepackage{booktabs}
\usepackage{multirow}
\usepackage{multicol}
\usepackage{makecell}
\usepackage[many]{tcolorbox}
\usepackage[table]{xcolor}

\newcommand{\lm}[1]{\texttt{#1}}

\newcommand{\data}[1]{\textsf{#1}}

\newtcolorbox{boxG}{
    fontupper = \color{black},
    rounded corners,
    arc = 6pt,
    colback = green!5!white, 
    colframe = green!45!black, 
    boxrule = 0pt, 
    bottomrule = 4.5pt,
    enhanced,
    fuzzy shadow = {0pt}{-3pt}{-0.5pt}{0.5pt}{black!35}
}

\newtcolorbox{boxB}{
    fontupper = \color{black},
    rounded corners,
    arc = 6pt,
    colback = blue!5!white, 
    colframe = blue!45!black, 
    boxrule = 0pt, 
    bottomrule = 4.5pt,
    enhanced,
    fuzzy shadow = {0pt}{-3pt}{-0.5pt}{0.5pt}{black!35}
}

\begin{document}
\maketitle
\begin{abstract}

The ability of large language models (LLMs) to process and generate text has introduced potential for applications in information extraction (IE). 
While it’s debated whether LLMs outperform smaller fine-tuned models for classification tasks, their strong generalization capability makes them promising for domains with limited labeled data available for fine-tuning. This advantage is particularly relevant for the emerging application of the digital product passport (DPP), where the problem space is broad but domain-specific data remains scarce.
Motivated by this use case, we apply generative IE to the product domain, explicitly addressing efficiency, generalizability, and data privacy constraints.
We propose a two-step validation method that integrates a PLM block into the generative IE pipeline and thereby leverages LLMs' correction capability.
We discover that such a validation task enhances LLM performance, particularly on the extraction of weakly expressed, low-salience entities that appear sparsely throughout the text.
For certain entities, the performance of mid-size models can even reach levels comparable to larger models, and the improvement of first-step PLM predictions also enhance the final LLM output.
Nevertheless, the effects on the smallest open-source LLMs (e.g., \lm{Llama-3.2 3B}) is limited.
Based on the findings, we develop a demo application for product information extraction that utilizes locally deployed LLMs, targeting further adaptations to real-world DPP use cases.\footnote{The repo for experiment is available at \url{https://github.com/doyouwantsometea/pie_paper}. The demo app is available at \url{https://github.com/hrw-neurolab/transferhub_pie_demo}.}

\end{abstract}

\section{Introduction}

As a long-established NLP task, information extraction (IE) remains challenging owing to the various and continually changing sources of information and task requirements \cite{DBLP:Xu2024}. Earlier works perform IE through fine-tuning pretrained language models (PLMs) such as \lm{BERT} \cite{ACL:jiang2020, ACL:Shin2020}, while more recent studies attempt to leverage the generation capability of LLMs to directly output structured data \cite{DBLP:Zhang2025}. Both mainstream approaches have their advantages and drawbacks: Fine-tuned PLMs can often surpass LLM performance owing to the task's classification-oriented nature \cite{DBLP:Wang2023}. By contrast, LLMs depend less on domain-specific data and generalize better than PLMs when such training data are scarce. In real-world use cases, the trade-off often imposes challenges to deploying robust IE applications that are both effective and well-integrated with domain-specific knowledge \cite{DBLP:Ma2023, DBLP:Blume2023, DBLP:Kim2024}.


\begin{figure}
    \centering
    \includegraphics[width=0.87\columnwidth]{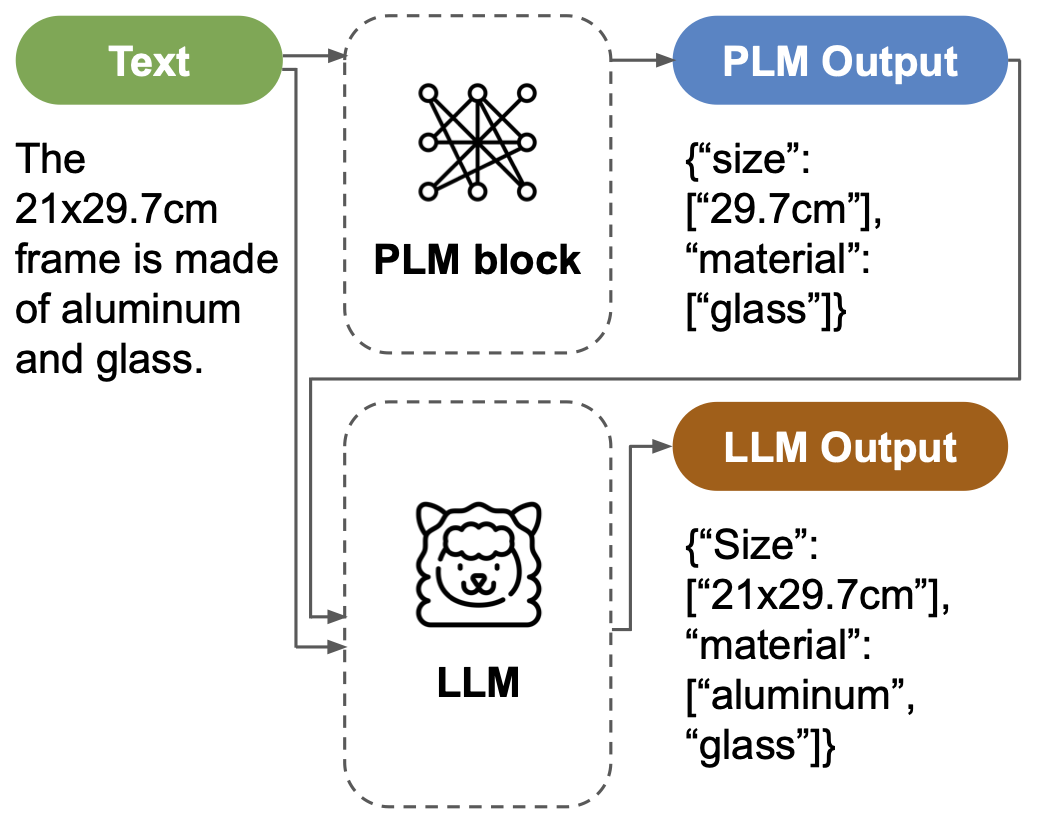}
    \caption{The workflow of the two-step validation task, where LLM is instructed to correct a PLM output instead of extracting information directly from the text.}
    \label{fig:workflow}
\end{figure}

This challenge is especially relevant to the product domain \cite{DBLP:Schon2018, DBLP:Brinkmann2024, DBLP:Hatty2024}: Structured, domain-specific annotations are costly and limited, while product information is highly heterogeneous and frequently subject to confidentiality constraints. These challenges draw our attention to this domain, particularly under the context of the European Commission's Digital Product Passport (DPP) \cite{DBLP:Petrik2025}, which requires the structured disclosure of product attributes including materials, components, and manufacturing details.
The implementation of the DPP is planned from 2027, starting from batteries and gradually extending to a broader coverage.
Although the establishment of the DPP can largely benefit from IE applications in creating and processing the required structured data, non-existing data and the broad scope introduce bottlenecks to addressing such a use case. Furthermore, while handling product specifications or confidential information, companies often avoid commercial LLMs (e.g. ChatGPT or Gemini) owing to data privacy concerns, hindering both the generation and use of structured data required by these new regulations. 

Motivated by the DPP use case, we focus on product information extraction using open-source LLMs in this study, eventually building a demo application that can be further extended to broaden use cases. Aiming at enhancing generalizability across underexplored product domains while reducing model size to allow local deployment, we reformulate the conventional generative IE approach as a validation task and instruct LLMs to correct first-step predictions from a fine-tuned PLM (Figure~\ref{fig:workflow}).
The twist strategically exploits LLMs’ text-refinement strengths to boost output quality without increasing model size, thereby achieving a more efficient performance-to-size ratio.
Our study contributes the following:
\begin{itemize}
    \item We introduce a hybrid, two-step generative IE method, bringing forward the validation task by integrating a PLM block into the generative IE pipeline.
    \item Our method enhances generative IE performance particularly on low-salience, weakly expressed entity classes. This enables smaller LLMs to deliver performance comparable to larger models and thereby benefits local deployment and data privacy.
    \item Based on the proposed validation task reformulation, we prototype a demo application (\S~\ref{sec:demo}) that can be further integrated to real-world applications, particularly under the context of the DPP.
\end{itemize}

\begin{table}
    \centering
    \small
    \begin{tabular}{lcc}
        \toprule
        \textbf{Entity} & \data{Amazon} & \data{E-commerce} \\
        \midrule
        Size & 570 & 349 \\
        Weight & 907 & 71 \\
        Product number & 710 & 133 \\
        Component & 993 & 718 \\
        Material & 637 & 400 \\
        Manufacturer & 722 & 447 \\
        \midrule
        \textbf{Data Instances} & 426 & 512 \\
        \bottomrule
    \end{tabular}
    \caption{Dataset size, entity classes and their counts in each dataset. While the first three entities are more explicit, the later ones tend to be weakly expressed in the text and are therefore considered more challenging.}
    \label{tab:entity}
\end{table}

\section{Related Work}

\paragraph{Conventional and generative IE.}

Information extraction (IE) transforms plain text into structured information and is commonly categorized into named entity recognition (NER), relation extraction (RE), and event extraction (EE) \cite{DBLP:Lu2022, DBLP:Xu2024}.
Conventional approaches for NER include \lm{BiLSTM-CRF} \cite{DBLP:Lample2016} and later on PLMs such as \lm{BERT} \cite{DBLP:BERT} and \lm{RoBERTa} \cite{DBLP:RoBERTa}; following the recent development of LLMs, their widespread use has shifted the task from its original classification-based nature to a generative one \cite{DBLP:Hsu2024, DBLP:Xu2024, DBLP:Zhang2025}.

Whether LLMs' generative capability suffices for IE tasks nevertheless remains a debatable topic. LLMs were often found outperformed by fine-tuned PLMs on IE tasks \cite{DBLP:Gao2024, DBLP:Peng2024}. Several studies highlighted the notable gap remaining between PLMs and LLMs \cite{DBLP:Han2023, DBLP:Li2023_ChatGPT, DBLP:RChen2024, DBLP:Liao2025}; LLMs were not considered sufficient on few-shot IE, with their performance possibly worsening on easy samples owing to hallucination or span boundary mismatch \cite{DBLP:Ma2023}. \citet{DBLP:Han2023} highlighted existing challenges LLMs, such as extending span lengths and being oversensitive to irrelevant context. According to \citet{DBLP:Wang2023}, LLMs without fine-tuning tended to fall behind supervised BERT-based models on NER task and may further suffer from known challenges such as hallucination.

\paragraph{Enhancing performance.}
Previous studies explored several approaches in pursuit of increasing the performance of generative IE. Generally, LLMs have been known to be capable of revising self-generated outputs for better results \cite{DBLP:Madaan2023, DBLP:Kamoi2024}.
When it comes to IE tasks, fine-tuning with fewer than 1k data instances could boost performance \cite{DBLP:Dunn2024}, and LLMs could also improve in NER when provided with self-annotated entities \cite{DBLP:Xie2024}. Alternatively, modifying output structure \cite{DBLP:Sainz2024} such as applying code-styled formulation \cite{DBLP:Li2025MPL} has been found beneficial.

LLMs were also often applied as a component in an IE workflow to achieve better overall performance. \citet{DBLP:Wei2023} proposed a two-step QA process to further break down zero-shot IE task. Similarly, \citet{DBLP:Li2024_GO} split the NER task into identifying named entities and formatting structured output. LLMs may also be added upon smaller models to perform post-hoc verification \cite{DBLP:Kim2024}, classification \cite{DBLP:Zhang2024}, or selection \cite{DBLP:WChen2024}. As \citet{DBLP:Zhang2025} highlighted the increased cost and computational demand with LLMs, \citet{DBLP:Kim2025} introduced a plug-in architecture that utilized both PLM and LLM in a pipeline to take into account efficiency and generalizability at once.

\paragraph{IE in the product domain.}

Although IE tasks are widely applicable to a range of real-world scenarios, when it comes to product information, the scarcity of training data often hinders the usage of NER applications with conventional methods \cite{DBLP:Schon2018}.
Nevertheless, the development of LLMs introduced a breakthrough to such a bottleneck. \citet{DBLP:Hatty2024} highlighted LLMs' outstanding capability of extracting explicit labels; moreover, LLMs could extract implicit labels, which was hardly achievable with the conventional token classification approach.
\citet{DBLP:Brinkmann2024} investigated product IE using GPT models and hinted at the potential of normalizing numeric attribute values. Similarly, \citet{DBLP:Li2025AM} probed into attribute mining in product description, introducing a framework adopting the chain-of-thought reasoning method \cite{DBLP:Wei2022}.





\section{Generative IE with Validation Task}
\label{sec:experiments}

\subsection{Task reformulation}

In our experiments, we adapt the conventional generative IE into a validation task, running LLMs on two task formulations:

\begin{enumerate}
    \item \textbf{\textit{Extraction}}, where the model is asked to identify entities from plain text and directly generate structured output.
    \item \textbf{\textit{Validation}}, which features two-step generation: An initial structured prediction is first made by the PLM. The prediction is then provided to the LLM, which validates and corrects the initial output (Figure~\ref{fig:workflow}).
\end{enumerate}

The validation task imitates the self-refine framework \cite{DBLP:Madaan2023} without recursive prompting. The method also echoes the usage of LLMs for post-hoc processing \cite{DBLP:Kim2024, DBLP:WChen2024} in NER workflow to leverage LLMs' general knowledge to the task.

In the validation task, we fine-tune one \lm{RoBERTa} and one \lm{DeBERTa} model per dataset, yielding four setups for first-step PLM predictions. As a baseline, LLMs are tasked with correcting an empty dictionary formatted as PLM output, adding up to a total of five setups for the validation task.

\subsection{Experiments}

\paragraph{Data.}

Focusing on product domain for further DPP applications, we use two open-source datasets as the main materials: Hugging Face \data{Amazon Product Description}\footnote{\url{https://huggingface.co/datasets/philschmid/amazon-product-descriptions-vlm}} (hereinafter \data{Amazon}/\data{AZ}) and Kaggle \data{E-commerce Text Classification}\footnote{\url{https://www.kaggle.com/datasets/saurabhshahane/ecommerce-text-classification}} (hereinafter \data{E-commerce}/\data{EC}). Both datasets are resampled to roughly 500 instances and then annotated with six named-entity labels (Table~\ref{tab:entity}) by a single annotator using \textsc{Label Studio} \cite{Label-Studio}. The definition of the labels is provided in Table~\ref{tab:guideline} in the Appendix.



\begin{table*}[h]
    \centering
    \small
    \begin{tabular}{clcccccc}
        \toprule
         & \multirow{2.5}{*}{\textbf{Model}} & \multirow{2.5}{*}{\textbf{\textit{Extraction}}} & \multicolumn{5}{c}{\textbf{\textit{Validation}}} \\
        \cmidrule{4-8}
         & & & \textit{baseline} & \lm{RoBERTa}-\data{AZ} & \lm{RoBERTa}-\data{EC} & \lm{DeBERTa}-\data{AZ} & \lm{DeBERTa}-\data{EC} \\
        \midrule
        \parbox[t]{2mm}{\multirow{7}{*}{\rotatebox[origin=c]{90}{\data{Amazon}}}} & \lm{Llama-3.2 3B} & 53.07 & \textbf{53.20} & 46.49 & 43.76 & 46.81 & 44.46 \\
         & \lm{Llama-3.1 8B} & 54.17 & 58.91 & \cellcolor{green!10}\textbf{61.45}* & \cellcolor{green!10}58.06 & \cellcolor{green!10}60.90 & \cellcolor{green!10}58.28 \\
         & \lm{Llama-3.3 70B} & 70.01 & 70.71 & \cellcolor{green!10}\textbf{72.94} & \cellcolor{green!10}71.44 & \cellcolor{green!10}72.84 & \cellcolor{green!10}71.33 \\
         & \lm{Mistral-0.3 7B} & 46.48 & \textbf{50.68} & 39.13* & 36.86 & 41.32 & 38.66 \\
         & \lm{Mistral-small-3.1 24B} & 66.49 & 62.89 & \cellcolor{green!10}69.36 & \cellcolor{green!10}67.20 & \cellcolor{green!10}\textbf{69.72} & \cellcolor{yellow!15}66.45 \\
         & \lm{Gemma-3 4B} & 53.24 & 52.09 & \cellcolor{green!10}55.19 & 50.88 &\cellcolor{green!10}\textbf{56.69} & \cellcolor{yellow!15}52.87 \\
         & \lm{Gemma-3 27B} & 65.03 & 66.50 & \cellcolor{green!10}\textbf{72.64} & \cellcolor{green!10}67.63 & \cellcolor{green!10}72.39 & \cellcolor{green!10}67.74 \\
        \midrule
        \parbox[t]{2mm}{\multirow{7}{*}{\rotatebox[origin=c]{90}{\data{E-commerce}}}} & \lm{Llama-3.2 3B} & \textbf{30.27} & 27.16 & 23.82 & \cellcolor{yellow!15}29.39 & 26.12 & \cellcolor{yellow!15}30.04 \\
         & \lm{Llama-3.1 8B} & 35.79 & 38.75 & \cellcolor{yellow!15}37.12 & \cellcolor{green!10}42.24* & \cellcolor{yellow!15}37.50 & \cellcolor{green!10}\textbf{42.80} \\
         & \lm{Llama-3.3 70B} & 44.48 & 44.89 & \cellcolor{green!10}46.32 & \cellcolor{green!10}46.49 & \cellcolor{green!10}46.03 & \cellcolor{green!10}\textbf{46.74} \\
         & \lm{Mistral-0.3 7B} & 30.48 & 33.08 & \cellcolor{yellow!15}32.74 & \cellcolor{green!10}34.75* & \cellcolor{green!10}33.52 & \cellcolor{green!10}\textbf{34.99} \\
         & \lm{Mistral-small-3.1 24B} & 44.57 & 43.23 & \cellcolor{yellow!15}44.23 & \cellcolor{green!10}49.64 & \cellcolor{green!10}46.57 & \cellcolor{green!10}\textbf{50.19} \\
         & \lm{Gemma-3 4B} & 37.54 & 37.89 & 37.50 & \cellcolor{green!10}41.35 & \cellcolor{yellow!15}37.62 & \cellcolor{green!10}\textbf{41.89} \\
         & \lm{Gemma-3 27B} & 46.20 & 45.52 & 45.50 & \cellcolor{green!10}\textbf{49.20} & \cellcolor{yellow!15}45.63 & \cellcolor{green!10}48.67 \\
        \bottomrule
    \end{tabular}
    \caption{F$_1$ score (\%) of generative IE across two task variations and different validation setups. The highest score per model is highlighted in bold face. Scores higher than both the extraction task and the validation baseline are marked in green, and those higher than either of them are in yellow. The starred scores denote the average over five runs to assess robustness.}
    \label{tab:llm_f1}
\end{table*}


\paragraph{Models.}

We include seven LLMs from three model families: \lm{Llama}, \lm{Mistral}, and \lm{Gemma} with different sizes, using open-source models to support local deployment for the DPP use cases. All the models are instruction-tuned variants downloaded from Hugging Face. The task is conducted through one-shot prompting with few-shot training. The prompt is provided in Figure~\ref{fig:prompt} in the Appendix.


\paragraph{Evaluation.} The outcome of generative IE is assessed through the F$_1$ score. Considering that gold labels might contain heterogeneous formats of a label (e.g. ``120cm'' and ``120 cm'' should be considered identical) and that LLMs tend to correct minor grammatical errors, we first eliminate spaces and special characters in both gold labels and predictions, and then remove duplicated entities and cases. The evaluation is conducted on an entity-level: A predicted entity is considered correct only while fully mapping a gold label of the same class. For example, ``wooden frame'' as a \textit{component} would be considered false because the gold labels mark ``frame'' as a \textit{component} and ``wood'' as a \textit{material}.

\section{Results and Discussion}





\paragraph{IE performance using validation method.}
We explore two generative IE task formulations and report the F$_1$ score in Table~\ref{tab:llm_f1}. To assess the robustness of the results, we repeated experiments with \lm{Llama-3.1 8B} and \lm{Mistral-0.3 7B}, together with fine-tuned \lm{RoBERTa} on both datasets, across five independent runs. The standard deviations range from $5.7 \times 10^{-3}$ to $1.2 \times 10^{-2}$, while the 95\% confidence intervals fall between $7.1 \times 10^{-3}$ and $1.5 \times 10^{-2}$. The low variance across different runs suggests that the observed performance remains stable.

Through the validation task, larger LLMs from the same model family tend to deliver better results, while the three model families do not substantially differ in performance. All models tend to score higher on \data{Amazon} across all settings; this may result from companies complying with certain format requirements in providing information on the Amazon website. Such a reason also leads to PLMs' outstanding capability of extracting \textit{size}, \textit{weight}, and \textit{product number} on the \data{Amazon} dataset (Appendix~\ref{sec:plm}).

While applying the two-step validation method, we find that mid-size and large models frequently benefit from the task reformulation, which echoes LLMs' capability of self-improvement \cite{DBLP:Madaan2023}. Most models score better F$_1$ in at least one scenario, with the improvement being the most consistent with \lm{Llama-3.3 70B}. Notably, even empty PLM predictions i.e. the baseline setup can already improve IE performance in multiple cases. This suggests that the gains are not solely driven by the good PLM predictions; since LLMs are proven capable of revising empty predictions, error propagation from incorrect PLM predictions to the final LLM output also becomes limited.
However, the smallest LLMs such as \lm{Llama-3.2 3B} and \lm{Gemma-3 4B} often suffer from a drastic performance drop, potentially because the longer prompted context increases difficulty of the task.

Precision and recall scores provide further details on how the validation task affects model performance. Although the smallest models could already achieve good precision with the intuitive extraction task, they suffer from significantly lower recall, which can be improved through the validation task. In comparison, the enhancement of precision score contributes more to the overall improvement with mid-sized and larger LLMs. Such an observation suggests that the two-step method could take advantages of LLMs' generalizability across models and validation settings in reducing ignored, undetected labels.

Among the four tested PLM blocks, predictions from a PLM fine-tuned on the same dataset consistently outperform other setups, boosting performance by up to over 7\% (\lm{Gemma-3 27B} on \data{Amazon}). Although the baseline setting is found already beneficial for the two-step method, this trend highlights the importance of PLM prediction quality, indicating that LLMs can directly take advantages of the improvements on the first-step prediction and thereby generate better final output.

\begin{figure}
    \centering
    \includegraphics[width=0.95\linewidth]{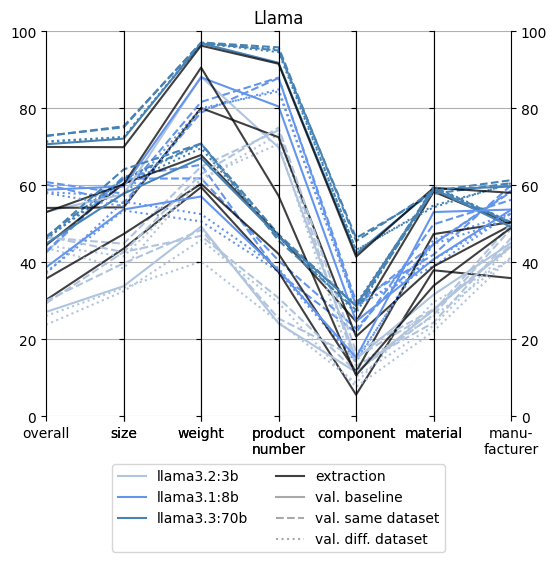}
    \includegraphics[width=0.95\linewidth]{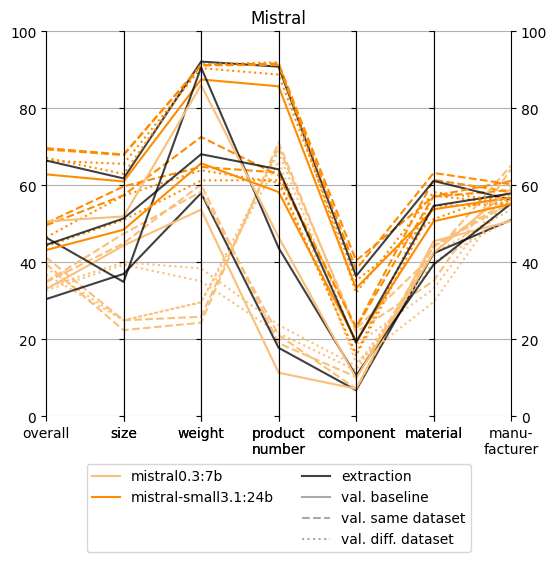}
    \includegraphics[width=0.95\linewidth]{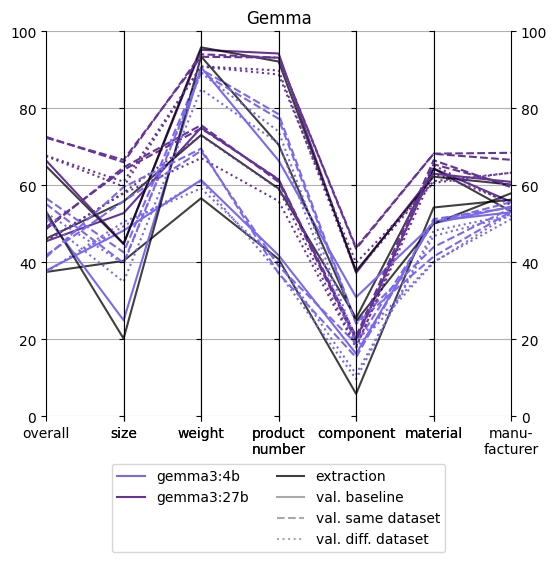}
    \caption{F$_1$ score (\%) per label for \lm{Llama} (top), \lm{Mistral} (middle), and \lm{Gemma} (bottom) model family. The black solid lines denote the extraction task, while the other line styles represent different validation setups.}
    \label{fig:f1}
\end{figure}

\paragraph{Enhancement on weakly expressed labels.} Although the overall F$_1$ score does not fluctuate massively, performance varies across labels. Figure~\ref{fig:f1} reveals that task reformulation introduces improvement more notably to \textit{component} and \textit{manufacturer}, whereas the original extraction task outperforms most validation setups on \textit{size} and \textit{weight}. This pattern holds across all three model families, with improvements reflected in both precision and recall (Appendix~\ref{sec:more_results}). Furthermore, on labels such as \textit{component} and \textit{manufacturer}, applying PLM block fine-tuned on the same dataset could enhance the performance of smaller models, enabling them to approximate and even occasionally outperform larger models.

This discrepancy highlights the advantages of the validation method on labels that are vague or appear sparsely in the text. Applications of product IE such as recommendation, decision-making \cite{DBLP:Brinkmann2024}, and the DPP \cite{DBLP:Petrik2025} can often depend on the weakly expressed labels. Moreover, the enhancement of lightweight LLMs enables local deployment even on limited hardware, making the method practical in use cases that have to comply with data privacy requirements. Overall, by reformulating generative IE and integrating the validation method, the approach offers strong potential for broader adoption and practical application in real-world scenarios.

\section{Demo Application}
\label{sec:demo}

We build a demo application upon the findings from the experiments to performs product information extraction, ultimately aiming to accelerate or even automate the process of implementing the DPP. The application adopts our proposed two-step method to generate structured information out of product descriptions in plain text, which we assume companies maintain as part of their product launch processes in the modern market. Considering the structured information demanded by the DPP remains unclear under the ongoing legislative process, we currently stick to the labels explored in the experiments for the application.

\begin{figure}
    \centering
    \includegraphics[width=\linewidth]{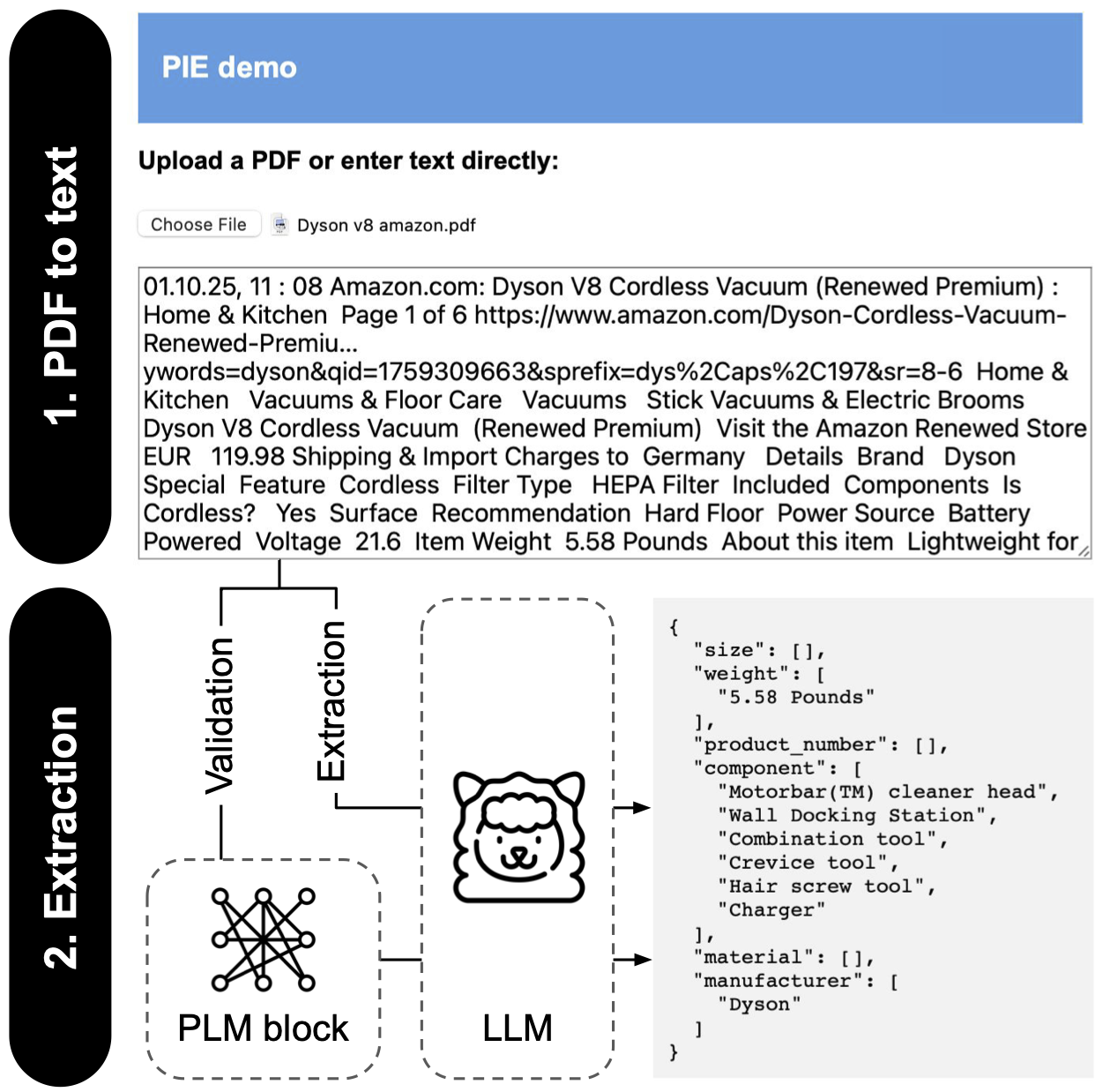}
    \caption{The architecture of the product information extractor demo tool, using the screenshot of processing an Amazon page of a vacuum cleaner as an example. In the demo, we split the labels into two groups: explicit and non-explicit, running both extraction and validation tasks in parallel to deliver better performance respectively on each group.}
    \label{fig:demo}
\end{figure}

Using examples from a web UI screenshot, Figure~\ref{fig:demo} visualizes the architecture of the application. In light of the divergent enhancement across the labels, we categorize the six labels into the explicit entities (\textit{size}, \textit{weight}, \textit{product number}) and the weakly expressed ones (\textit{component}, \textit{material}, \textit{manufacturer}), assigning the two groups with different tasks: The explicit entities are directly extracted by the LLM, while the weakly expressed entities undergo the two-step validation process. Among the labels, we consider \textit{component} and \textit{material} the most relevant to a DPP use case. For the validation task, we use fine-tuned \lm{RoBERTa} as the PLM block despite its slightly poorer performance in comparison to \lm{DeBERTa}, since the inference time is significantly shorter. Based on the respectively better performing method, the label split enhances the overall extraction quality.

Although the F$_1$ scores (Figure~\ref{fig:f1}) suggest that the smallest LLMs (e.g., \lm{Llama-3.2 3B} and \lm{Gemma-3 4B}) suffer from performance drop under the validation task, the small size makes them to some extent usable while facing strict hardware constraints, particularly considering the increased recall: Extraction task typically yield lower recall for weakly expressed labels, especially \textit{component}. From this point of view, improvements are already noticeable with even the smallest LLMs (Figure~\ref{fig:recall}). Although the task reformulation has limited impact on precision, it helps reduce false negative predictions, which is already beneficial in our use case. A possible minor factor is the task split reduces the label amount from 6 to 3, which potentially simplifies the task with one-shot prompting. With the smallest LLMs, the application can be efficiently deployed on a local device without the need for a traditional GPU. For example, in one of our tested environments with an M1 MacBookPro (produced 2021), processing a one-page-long product description usually takes less than 5 seconds with \lm{Llama-3.2}, highlighting the advantage in lowering processing time.

The application provides companies, especially small and medium-sized enterprises (SMEs), a solution to utilize existing data (e.g. product description) to simplify the establishment of DPP while conforming to the upcoming EU regulations. Leveraging the validation task, the application improves the performance-to-size ratio of generative IE workflow and provides more flexibility with local setups. This also prevents data leakage and guarantees privacy, which is a huge concern for SMEs to process sensitive data with commercial models such as GhatGPT. Through further adaptation to the updated regulations, the application demonstrates the potential for efficient deployment and integration into other systems.

\section{Conclusion}

In this study, we examined generative IE methods in the product domain to support applications regarding the DPP.
We introduced a two-step validation method to reformulate generative IE task and discovered that open-source LLMs can enhance IE performance in most cases.
Furthermore, the F$_1$ score enhancement was proportional to the quality of PLM predictions given as the source to correct.
The improvement was most evident for weakly expressed entities with higher semantic complexity; more explicitly stated entities were already well handled by standard extraction methods.
Based on the findings, we further implemented a demo application that performs IE on product description, aiming to accelerate the establishment of the DPP.
The enhancement of lightweight LLMs in the IE pipeline enables local deployment, ensuring stronger data privacy for companies and eventually paving the way for broader real-world adoption.

\section*{Limitations}

The datasets are annotated by only one annotator because of the dataset size and the clear entity span; although the labels are straightforward as product information tend to prevent semantic ambiguity, mislabeling may occur and induce biases into the results.
The annotated spans can sometimes be affected by tokenization. In particular, we find the mixture of alphabet and numbers (e.g. as ``1x1.2 m'' or ``ABC1000TUV'') occasionally incorrectly tokenized, which propagates to the following evaluation.
While evaluating LLM output, strict matching does not fully capture how practical the results are from an application-oriented perspective. When the gold label marks ``gloves'' as a component, ``white gloves'' would be considered false, whereas the later may seem slightly better considering the DPP use case.
Furthermore, we employ token-level F$_1$ to evaluate fine-tuned PLMs and yet entity-level F$_1$ for LLMs. It is therefore challenging to make direct comparisons between scores from the two approaches.

\section*{Ethical considerations}

Early stage of framing the research directions involve using ChatGPT (\lm{GPT-5.1}) to review feasibility of new ideas.
Regarding contents, we do not see immediate ethical concerns in terms of research and development.

\section*{Acknowledgments}

This work has been funded by the Ministry of Economy, Innovation, Digitalization, and Energy of the State of North Rhine-Westphalia, Germany, and the European Union within the project \textit{Der Transferhub Digitalisierung \& Circular Economy im Prosperkolleg}. We thank Nils Feldhus for his careful review and valuable feedback of the paper draft.


\bibliography{custom}

\appendix

\section{IE with PLMs}
\label{sec:plm}



\begin{table*}[h]
    \centering
    \tiny
    \begin{tabular}{clccccccccccccccc}
        \toprule
        \multirow{2.5}{*}{\textbf{Data}} & \multirow{2.5}{*}{\textbf{Model}} & \multirow{2.5}{*}{\textbf{O}} & \multicolumn{2}{c}{\textit{Size}} & \multicolumn{2}{c}{\textit{Weight}} & \multicolumn{2}{c}{\textit{PN}} & \multicolumn{2}{c}{\textit{Component}} & \multicolumn{2}{c}{\textit{Material}} & \multicolumn{2}{c}{\textit{Manufacturer}} & \multirow{2.5}{*}{\textit{Overall}} & \multirow{2.5}{*}{\textit{\makecell{Cross \\ eval}}}\\
        \cmidrule(l{2pt}r{2pt}){4-5} \cmidrule(l{2pt}r{2pt}){6-7} \cmidrule(l{2pt}r{2pt}){8-9} \cmidrule(l{2pt}r{2pt}){10-11} \cmidrule(l{2pt}r{2pt}){12-13} \cmidrule(l{2pt}r{2pt}){14-15}
         & & & \textbf{B} & \textbf{I} & \textbf{B} & \textbf{I} & \textbf{B} & \textbf{I} & \textbf{B} & \textbf{I} & \textbf{B} & \textbf{I} & \textbf{B} & \textbf{I} & & \\
        \midrule
        \multirow{2}{*}{\data{AZ}} & \lm{RoBERTa} & \textbf{98.40} & 90.32 & 91.58 & 96.00 & 95.26 & 93.76 & 93.18 & 54.60 & 52.11 & 75.93 & \textbf{77.17} & \textbf{72.20} & 71.33 & \textbf{96.99} & 93.72 \\
         & \lm{DeBERTa} & 98.35 & \textbf{91.94} & \textbf{92.12} & \textbf{96.87} & \textbf{96.20} & \textbf{93.80} & \textbf{93.48} & \textbf{55.11} & 46.25 & \textbf{77.08} & 73.94 & 71.62 & \textbf{71.98} & 96.92 & 94.19 \\
        \multirow{2}{*}{\data{EC}} & \lm{RoBERTa} & 97.37 & 69.17 & 77.95 & 73.50 & 75.00 & 64.61 & 75.02 & 46.53 & \textbf{52.29} & 73.55 & 67.53 & 67.57 & 66.66 & 94.83 & 94.57 \\
         & \lm{DeBERTa} & 97.29 & 70.01 & 80.36 & 88.85 & 85.36 & 65.30 & 66.31 & 39.45 & 35.42 & 75.74 & 68.61 & 68.88 & 69.94 & 94.43 & \textbf{94.72} \\
        \bottomrule
    \end{tabular}
    \caption{F$_1$ score (\%) of fine-tuned PLMs on NER under the BIO schema. The overall score is weighted by the number of tokens. The scores denote the mean value of the three runs, and the highest score per label is highlighted in bold face.}
    \label{tab:plm_f1}
\end{table*}

\begin{table*}[h]
    \centering
    \tiny
    \begin{tabular}{clcccccccccccccc}
        \toprule
        \multirow{2.5}{*}{\textbf{Data}} & \multirow{2.5}{*}{\textbf{Model}} & \multirow{2.5}{*}{\textbf{O}} & \multicolumn{2}{c}{\textit{Size}} & \multicolumn{2}{c}{\textit{Weight}} & \multicolumn{2}{c}{\textit{PN}} & \multicolumn{2}{c}{\textit{Component}} & \multicolumn{2}{c}{\textit{Material}} & \multicolumn{2}{c}{\textit{Manufacturer}} & \multirow{2.5}{*}{\textit{Overall}}\\
        \cmidrule(l{2pt}r{2pt}){4-5} \cmidrule(l{2pt}r{2pt}){6-7} \cmidrule(l{2pt}r{2pt}){8-9} \cmidrule(l{2pt}r{2pt}){10-11} \cmidrule(l{2pt}r{2pt}){12-13} \cmidrule(l{2pt}r{2pt}){14-15}
         & & & \textbf{B} & \textbf{I} & \textbf{B} & \textbf{I} & \textbf{B} & \textbf{I} & \textbf{B} & \textbf{I} & \textbf{B} & \textbf{I} & \textbf{B} & \textbf{I} & \\
        \midrule
        \multirow{2}{*}{\data{AZ}} & \lm{RoBERTa} & 96.91 & 56.81 & 72.73 & 38.38 & 35.36 & 38.20 & 57.14 & 23.05 & 22.12 & \textbf{62.80} & 54.14 & 26.67 & 41.32 & 93.72 \\
         & \lm{DeBERTa} & 97.13 & 60.44 & 75.59 & 52.92 & 48.80 & 58.44 & 66.13 & 25.87 & 24.80 & 62.51 & 50.92 & 42.69 & 43.34 & 94.19 \\
        \multirow{2}{*}{\data{EC}} & \lm{RoBERTa} & 97.30 & 72.54 & 85.00 & 87.10 & 89.69 & \textbf{80.76} & \textbf{87.93} & 30.86 & 24.10 & 57.72 & \textbf{54.74} & 41.94 & \textbf{48.30} & 94.57 \\
         & \lm{DeBERTa} & \textbf{97.38} & \textbf{78.14} & \textbf{87.61} & \textbf{95.37} & \textbf{96.03} & 58.15 & 82.52 & \textbf{36.02} & \textbf{28.75} & 59.29 & 55.13 & \textbf{45.64} & 48.26 & \textbf{94.72} \\
        \bottomrule
    \end{tabular}
    \caption{F$_1$ score (\%) of cross-evaluation under the BIO schema. The overall score is weighted by the number of tokens. The scores denote the mean value of the three runs, and the highest score per label is highlighted in bold face.}
    \label{tab:plm_cross_eval}
\end{table*}

\begin{table*}
    \centering
    \small
    \begin{tabular}{ll}
        \toprule
        \textbf{Label} & \textbf{Description} \\
        \midrule
        Size & Size measurement of the product, e.g. 120 cm \\
        Weight & Weight measurement of the product, e.g. 1 kg \\
        Product number & Numeric or alphabetic identifier of the product, e.g. AM12345678 \\
        Component & Product composition and removable segments, e.g. frame, cable, battery \\
        Material & Material of the product or its components, e.g. polyester \\
        Manufacturer & Name of the Manufacturer, e.g. IKEA\\
        \bottomrule
    \end{tabular}
    \caption{Labels and the corresponding descriptions provided as the annotation guideline.}
    \label{tab:guideline}
\end{table*}

We fine-tune BERT-based models as the PLM block in the two-step validation method. We first perform hyperparameter search with \lm{BERT} \cite{DBLP:BERT}, \lm{DistilBERT} \cite{DBLP:DistilBERT}, \lm{RoBERTa} \cite{DBLP:RoBERTa}, and \lm{DeBERTa} \cite{DBLP:DeBERTa}, deciding to use the later two because of better performance.

\begin{table*}[h]
    \centering
    \scriptsize
    \begin{tabular}{clcccccc}
        \toprule
         & \multirow{2.5}{*}{\textbf{Model}} & \multirow{2.5}{*}{\textbf{\textit{Extraction}}} & \multicolumn{5}{c}{\textbf{\textit{Validation}}} \\
        \cmidrule{4-8}
         & & & \textit{baseline} & \lm{RoBERTa}-\data{AZ} & \lm{RoBERTa}-\data{EC} & \lm{DeBERTa}-\data{AZ} & \lm{DeBERTa}-\data{EC} \\
        \midrule
        \parbox[t]{2mm}{\multirow{7}{*}{\rotatebox[origin=c]{90}{\data{Amazon}}}} & \lm{Llama-3.2 3B} & \textbf{67.64} & 61.11 & 49.15 & 48.06 & 49.54 & 46.16 \\
         & \lm{Llama-3.1 8B} & \textbf{66.96} & 64.97 & \cellcolor{yellow!15}66.87* & 62.32 & \cellcolor{yellow!15}66.32 & 63.13 \\
         & \lm{Llama-3.3 70B} & 63.96 & 66.04 & \cellcolor{green!10}\textbf{68.25} & \cellcolor{green!10}67.24 & \cellcolor{green!10}68.18 & \cellcolor{green!10}66.95 \\
         & \lm{Mistral-0.3 7B} & 61.87 & \textbf{65.26} & 43.31* & 41.54 & 45.39 & 42.05 \\
         & \lm{Mistral-small-3.1 24B} & 64.36 & 63.52 & \cellcolor{green!10}70.95 & \cellcolor{green!10}69.04 & \cellcolor{green!10}\textbf{71.60} & \cellcolor{green!10}68.79 \\
         & \lm{Gemma-3 4B} & \textbf{58.57} & 54.58 & \cellcolor{yellow!15}54.89 & 51.31 & \cellcolor{yellow!15}56.85 & 52.33 \\
         & \lm{Gemma-3 27B} & 61.44 & 65.88 & \cellcolor{green!10}\textbf{71.13} & \cellcolor{green!10}66.70 & \cellcolor{green!10}70.95 & \cellcolor{green!10}65.90 \\
        \midrule
        \parbox[t]{2mm}{\multirow{7}{*}{\rotatebox[origin=c]{90}{\data{E-commerce}}}} & \lm{Llama-3.2 3B} & \textbf{34.96} & 23.83 & 21.24 & \cellcolor{yellow!15}25.74 & 22.99 & \cellcolor{yellow!15}26.61 \\
         & \lm{Llama-3.1 8B} & 36.71 & 36.78 & 34.87 & \cellcolor{green!10}40.19* & 35.55 & \cellcolor{green!10}\textbf{40.38} \\
         & \lm{Llama-3.3 70B} & 37.02 & 36.57 & \cellcolor{green!10}38.62 & \cellcolor{green!10}38.84 & \cellcolor{green!10}38.50 & \cellcolor{green!10}\textbf{39.13} \\
         & \lm{Mistral-0.3 7B} & 29.30 & 31.48 & \cellcolor{yellow!15}30.31 & \cellcolor{green!10}32.72* & \cellcolor{green!10}31.77 & \cellcolor{green!10}\textbf{32.77} \\
         & \lm{Mistral-small-3.1 24B} & 43.77 & 43.26 & \cellcolor{green!10}44.91 & \cellcolor{green!10}50.38 & \cellcolor{green!10}47.53 & \cellcolor{green!10}\textbf{50.84} \\
         & \lm{Gemma-3 4B} & \textbf{42.97} & 34.24 & \cellcolor{yellow!15}35.08 & \cellcolor{yellow!15}38.09 & \cellcolor{yellow!15}35.15 & \cellcolor{yellow!15}38.37 \\
         & \lm{Gemma-3 27B} & 39.63 & 41.92 & \cellcolor{yellow!15}41.72 & \cellcolor{green!10}\textbf{45.75} & \cellcolor{green!10}42.47 & \cellcolor{green!10}45.03 \\
        \bottomrule
    \end{tabular}
    \caption{Precision score (\%) of generative IE across two task variations and different validation setups. The highest score per model is highlighted in bold face. Scores higher than both the extraction task and the validation baseline are marked in green, and those higher than either of them are in yellow. The starred numbers denote the averages over five runs. Across the four reported mean scores, standard deviations range from $4.1 \times 10^{-3}$ to $1.0 \times 10^{-2}$, while 95\% confidence intervals fall between $5.1 \times 10^{-3}$ and $1.3 \times 10^{-2}$.}
    \label{tab:llm_precision}
\end{table*}

\begin{table*}[h]
    \centering
    \scriptsize
    \begin{tabular}{clcccccc}
        \toprule
         & \multirow{2.5}{*}{\textbf{Model}} & \multirow{2.5}{*}{\textbf{\textit{Extraction}}} & \multicolumn{5}{c}{\textbf{\textit{Validation}}} \\
        \cmidrule{4-8}
         & & & \textit{baseline} & \lm{RoBERTa}-\data{AZ} & \lm{RoBERTa}-\data{EC} & \lm{DeBERTa}-\data{AZ} & \lm{DeBERTa}-\data{EC} \\
        \midrule
        \parbox[t]{2mm}{\multirow{7}{*}{\rotatebox[origin=c]{90}{\data{Amazon}}}} & \lm{Llama-3.2 3B} & 43.67 & \textbf{47.11} & \cellcolor{yellow!15}44.10 & 40.17 & \cellcolor{yellow!15}44.36 & 42.88 \\
         & \lm{Llama-3.1 8B} & 45.49 & 53.88 & \cellcolor{green!10}\textbf{56.86}* & \cellcolor{green!10}54.35 & \cellcolor{green!10}56.30 & \cellcolor{green!10}54.12 \\
         & \lm{Llama-3.3 70B} & 77.32 & 76.10 & \cellcolor{green!10}\textbf{78.31} & \cellcolor{yellow!15}76.20 & \cellcolor{green!10}78.18 & \cellcolor{yellow!15}76.33 \\
         & \lm{Mistral-0.3 7B} & 37.22 & \textbf{41.42} & 35.69* & 33.12 & \cellcolor{yellow!15}37.92 & 35.77 \\
         & \lm{Mistral-small-3.1 24B} & \textbf{68.76} & 62.28 & \cellcolor{yellow!15}67.83 & \cellcolor{yellow!15}65.45 & \cellcolor{yellow!15}67.93 & \cellcolor{yellow!15}64.26 \\
         & \lm{Gemma-3 4B} & 48.79 & 49.82 & \cellcolor{green!10}55.50 & \cellcolor{green!10}50.45 & \cellcolor{green!10}\textbf{56.53} & \cellcolor{green!10}53.42 \\
         & \lm{Gemma-3 27B} & 69.06 & 61.74 & \cellcolor{green!10}\textbf{74.21} & \cellcolor{yellow!15}68.60 & \cellcolor{green!10}73.88 & \cellcolor{green!10}69.69 \\
        \midrule
        \parbox[t]{2mm}{\multirow{7}{*}{\rotatebox[origin=c]{90}{\data{E-commerce}}}} & \lm{Llama-3.2 3B} & 26.69 & 31.57 & \cellcolor{yellow!15}27.12 & \cellcolor{green!10}34.25 & 20.23 & \cellcolor{green!10}\textbf{34.49} \\
         & \lm{Llama-3.1 8B} & 34.92 & 40.95 & \cellcolor{yellow!15}39.67 & \cellcolor{green!10}44.54* & \cellcolor{yellow!15}39.67 & \cellcolor{green!10}\textbf{45.52} \\
         & \lm{Llama-3.3 70B} & 55.70 & \textbf{58.14} & \cellcolor{yellow!15}57.83 & \cellcolor{yellow!15}57.89 & \cellcolor{yellow!15}57.22 & \cellcolor{yellow!15}58.01 \\
         & \lm{Mistral-0.3 7B} & 31.75 & 34.86 & \cellcolor{green!10}35.59 & \cellcolor{green!10}37.04* & \cellcolor{green!10}35.47 & \cellcolor{green!10}\textbf{37.54} \\
         & \lm{Mistral-small-3.1 24B} & 45.40 & 43.21 & \cellcolor{yellow!15}43.57 & \cellcolor{green!10}48.93 & \cellcolor{green!10}45.64 & \cellcolor{green!10}\textbf{49.54} \\
         & \lm{Gemma-3 4B} & 33.33 & 42.41 & \cellcolor{yellow!15}40.28 & \cellcolor{green!10}45.22 & \cellcolor{yellow!15}40.46 & \cellcolor{green!10}\textbf{46.13} \\
         & \lm{Gemma-3 27B} & \textbf{55.39} & 49.79 & \cellcolor{yellow!15}50.03 & \cellcolor{yellow!15}53.20 & 49.30 & \cellcolor{yellow!15}52.96 \\
        \bottomrule
    \end{tabular}
    \caption{Reacll score (\%) of generative IE across two task variations and different validation setups. The highest score per model is highlighted in bold face. Scores higher than both the extraction task and the validation baseline are marked in green, and those higher than either of them are in yellow. The starred numbers denote the averages over five runs. Across the four reported mean scores, standard deviations range from $6.1 \times 10^{-3}$ to $1.8 \times 10^{-2}$, while the corresponding 95\% confidence intervals fall between $7.6 \times 10^{-3}$ and $2.3 \times 10^{-2}$.}
    \label{tab:llm_recall}
\end{table*}

\subsection{Experiments}

\paragraph{Setup.}
We annotate the datasets with six entity classes, using the label description as guidelines (Table~\ref{tab:guideline}). The same annotation is also used in the experiments (\S~\ref{sec:experiments}).
For each setup of dataset and PLM, we run the fine-tuning for three times, using 5-fold cross-validation and training for 8 epochs with a batch size of 16 and a learning rate set to $2 \times 10^{-5}$.
The experiments are conducted on 4 NVIDIA Quadro RTX8000 GPUs.



\paragraph{Model selection.} While performing the validation task reformulation, we select one \lm{RoBERTa} and one \lm{DeBERTa} per dataset, making four setups in total for generating the first-step PLM prediction. In fine-tuning, the repeated experiment runs produce consistent model performance, with the overall F$_1$ scores deviating by less than 1\%. We thereby further compare the models through evaluating them with the dataset they are \textit{not} fine-tuned on, selecting the model scoring the highest accuracy under this cross-evaluation schema, which implies better generalizability over unseen data.


\begin{figure}
    \centering
    \includegraphics[width=0.95\linewidth]{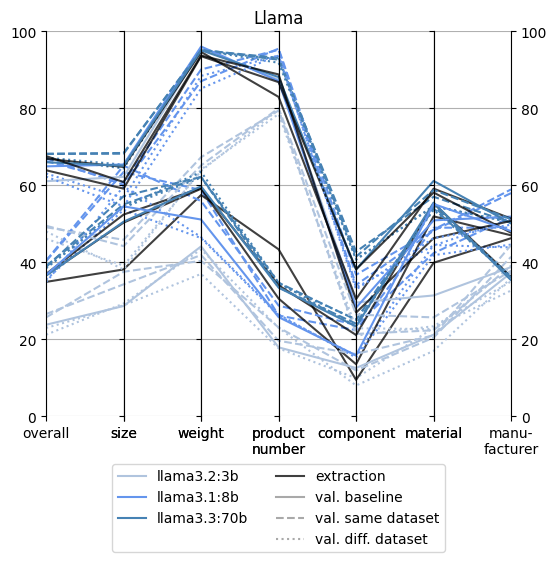}
    \includegraphics[width=0.95\linewidth]{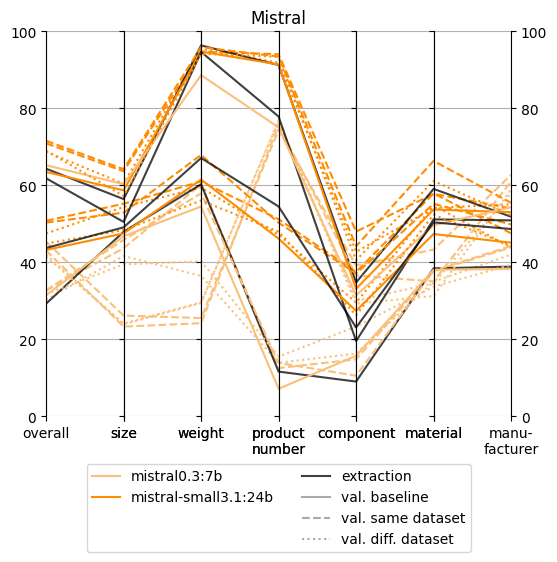}
    \includegraphics[width=0.95\linewidth]{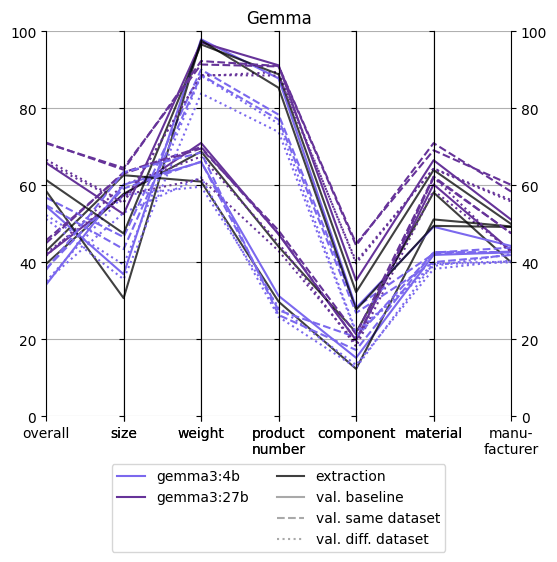}
    \caption{Precision score (\%) per label for \lm{Llama} (top), \lm{Mistral} (middle), and \lm{Gemma} (bottom) model family. The black solid lines denote the extraction task, while the other line styles represent different validation setups.}
    \label{fig:precision}
\end{figure}

\begin{figure}
    \centering
    \includegraphics[width=0.95\linewidth]{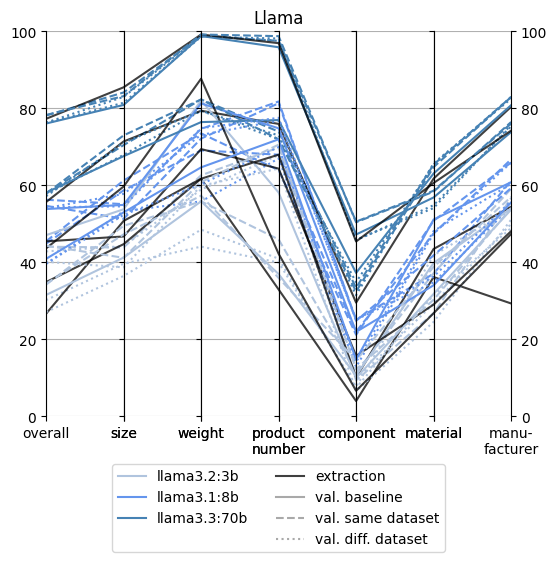}
    \includegraphics[width=0.95\linewidth]{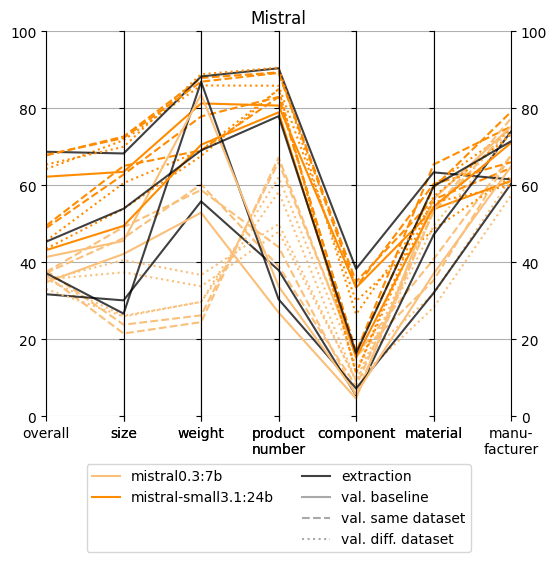}
    \includegraphics[width=0.95\linewidth]{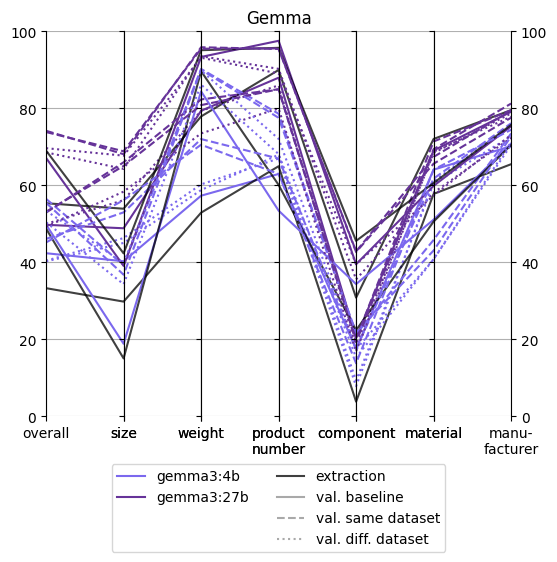}
    \caption{Recall score (\%) per label for \lm{Llama} (top), \lm{Mistral} (middle), and \lm{Gemma} (bottom) model family. The black solid lines denote the extraction task, while the other line styles represent different validation setups.}
    \label{fig:recall}
\end{figure}

\subsection{Results}

For each setting, we run the experiments three times and report the mean score in Table~\ref{tab:plm_f1}. Across the models and datasets, the standard deviations of overall F$_1$ range from $3.0 \times 10^{-4}$ to $3.7 \times 10^{-3}$, while the standard deviations of cross-evaluation from $9.7 \times 10^{-4}$ to $3.8 \times 10^{-3}$.

PLMs achieve high overall F$_1$ scores, though this is largely driven by performance on the O-label, masking struggles with specific entity types. We find that PLMs extract information more accurately at entities that are more explicitly stated in the text, such as \textit{weight} and \textit{product number}, as they tend to show up in specifications and could often include numeric values. In contrast, \textit{component} becomes the most challenging entity for both models since it involves more complex semantic comprehension over a paragraph or even the entire text.

Though performing cross-evaluation using a different dataset, we further examine fine-tuned PLMs' generalizability (Table~\ref{tab:plm_cross_eval}). While using a different dataset for evaluation, we find performance more significantly dropping on the weakly expressed entities such as \textit{component} and \textit{manufacturer}. The trend highlights data collection and labeling as bottlenecks in applying conventional methods to real-world IE tasks, as the model may still fail to generalize to unseen data instances in spite of the shared domain knowledge.

\section{Supplementary Results}
\label{sec:more_results}

Extending F$_1$ scores in Table~\ref{tab:llm_f1}, we report precision (Table~\ref{tab:llm_precision}) and recall (Table~\ref{tab:llm_recall}) scores of generative IE tasks and setups. For the scores on each entity class, we visualize the scores in Figure~\ref{fig:precision} and \ref{fig:recall}.

\begin{figure*}
    \centering
    \small
    \begin{boxG}
        You are an expert in information extraction. You are given a task to find information described in a free-text product description, and put them into structured form.\\
        The designated output format should be in JSON and contains these keys: \{"size":"", "weight": "", "product number": "", "composition": "", "material": "", "manufacturer": ""\}\\\\Follow these principles: \\1. Be concise and avoid verbose explanations.\\2. Use a list as value to include all the information. Each key may include zero, one, or multiple suitable values.\\3. Some information may be lacking in the input text. If no proper information could be retrieved, put None as the value.\\4. Output only the designated JSON, and don't add any keys to its format.\\\\
        Here are some examples of a free-text product description and the target output:\\
        **Input:** This \#VAGA-198964 table from Ikea is composed of a wooden surface and a steel frame. Package size: 2x2m | Package weight: 5.8kg | Shipping policy: Free delivery from 59€ \\ **Output:** \{"size": ["2x2m"], "weight": ["5.8kg"], "product number": ["\#VAGA-198964"], "composition": ["surface", "frame"], "material": ["wood", "steel"], "manufacturer": ["Ikea"]\}\\
        **Input:** The ergonomic wireless mouse is designed specifically for right-handed users. It is also pleasant to the touch and encourages a natural hand position. Thanks to the interference-free 2.4 GHz wireless technology, the mouse can be used flexibly within a range of up to 10 m - without any cables at all \\
        **Output:** \{"size": [], "weight": [], "product number": [], "composition": [], "material": [], "manufacturer": []\},\\
        **Input:** The bed suite from Zara Home features two pillows and one quilt. Suitable for double bed (140x200), they are made of cotton fibre and are machine washable. \\
        **Output:** \{"size": ["140x220"], "weight": [], "product number": [], "composition": ["pillow", "quilt"], "material": ["cotton"], "manufacturer": ["Zara Home"]\}\\\\
        Perform the given task on the following text: \{TEXT\}\\You must return valid JSON with the required keys at the top level without adding any text. If product description or any information is missing, return an empty list for that field. Do not return error messages or any other keys.
    \end{boxG}
    \begin{boxB}
        You are an expert in processing product data. Your task is to validate structured output from a language model on information extraction. Given the input and model output, check whether the model-generated answer is correct or should be revised.\\
        Stick to this JSON format and do not add any keys: \{"size":"", "weight": "", "product number": "", "composition": "", "material": "", "manufacturer": ""\}\\\\Follow these principles:\\1. Be concise and avoid verbose explanations.\\2. Use a list as value to include all the information. Each key may include zero, one, or multiple suitable values.\\3. Make minimal corrections to the model-generated answer. If it already looks correct, return it as the output.\\4. Do not include information that is not mentioned in the text.
        
        Here are some examples of a free-text product description and the validated, corrected target output:\\
        **Input:** This \#VAGA-198964 table from Ikea is composed of a wooden surface and a steel frame. Package size: 2x2m | Package weight: 5.8kg | Shipping policy: Free delivery from 59€ \\ **Output:** \{"size": ["2x2m"], "weight": ["5.8kg"], "product number": ["\#VAGA-198964"], "composition": ["surface", "frame"], "material": ["wood", "steel"], "manufacturer": ["Ikea"]\}\\
        **Input:** The ergonomic wireless mouse is designed specifically for right-handed users. It is also pleasant to the touch and encourages a natural hand position. Thanks to the interference-free 2.4 GHz wireless technology, the mouse can be used flexibly within a range of up to 10 m - without any cables at all \\
        **Output:** \{"size": [], "weight": [], "product number": [], "composition": [], "material": [], "manufacturer": []\},\\
        **Input:** The bed suite from Zara Home features two pillows and one quilt. Suitable for double bed (140x200), they are made of cotton fibre and are machine washable. \\
        **Output:** \{"size": ["140x220"], "weight": [], "product number": [], "composition": ["pillow", "quilt"], "material": ["cotton"], "manufacturer": ["Zara Home"]\}\\\\
        Perform the given task on the following text: \{TEXT\} \\You must return valid JSON with the required keys at the top level without adding any text. If product description or any information is missing, return an empty list for that field. Do not return error messages or any other keys.
    \end{boxB}
    \caption{Prompt structure of the extraction (green) and validation task (blue). Considering that the instances from the datasets feature long text and thereby sparse named entities, we provide three manually created few-shot examples in the prompt, each containing no more than three sentences to ensure high entity density. Throughout the experiments, we identify issues with LLM output such as additional keys and values, gradually optimizing the instructions to ensure the robustness of the prompt.}
    \label{fig:prompt}
\end{figure*}

\end{document}